\RequirePackage{silence}
\documentclass[runningheads]{llncs}

\usepackage[T1]{fontenc}
\IfFileExists{newtxtext.sty}{%
  \usepackage{newtxtext}%
  \usepackage[varvw]{newtxmath}%
}{\PackageWarning{main}{newtx not installed -- draft build only}}
\usepackage{graphicx}
\usepackage{booktabs}
\usepackage{microtype}
\usepackage{etoolbox}
\AtBeginEnvironment{thebibliography}{%
  \emergencystretch=2.5em
  \hbadness=10000
}
\usepackage{xcolor}
\usepackage{multirow}
\usepackage{float}
\usepackage{needspace}
\usepackage{placeins}
\usepackage[misc]{ifsym}

\newcommand{\runin}[1]{\par\noindent\textbf{#1}\ }
\makeatletter
\renewcommand\subsubsection{\@startsection{subsubsection}{3}{\z@}%
  {\z@ \@plus 0.2\p@ \@minus 0.2\p@}%
  {-0.5em \@plus -0.22em \@minus -0.1em}%
  {\normalfont\normalsize\bfseries\boldmath}}
\makeatother

\begin{document}
\raggedbottom

\title{HEmIF-Bench: Benchmarking H\&E-to-mIF Models Beyond Virtual-Stain Fidelity}
\titlerunning{HEmIF-Bench}

\author{Fuxian Zhao\inst{1}(\Letter) \and Hyokeun Park\inst{2}(\Letter)}
\authorrunning{F. Zhao and H. Park}

\institute{%
  Department of Physics, The Hong Kong University of Science and Technology, Hong Kong SAR, China\\
  \email{fzhaoai@connect.ust.hk} \and
  Division of Life Science, The Hong Kong University of Science and Technology, Hong Kong SAR, China \\
  \email{hkpark@ust.hk}
}
\maketitle


\begin{abstract}
Hematoxylin and eosin (H\&E)-to-multiplex immunofluorescence (mIF)
generators predict spatial protein expression from routine histology,
yet pixel-level marker reconstruction does not establish downstream
utility.
We present HEmIF-Bench, evaluating five frozen generators across eight
datasets spanning molecular, cellular, tissue, and patient scales.
HEmIF-Bench decouples internal encoder features from generated outputs,
benchmarking both against the pathology foundation reference UNI2.
Mirroring the clinical workflow where pathologists rely on mIF to define
cell phenotypes, delineate tertiary lymphoid structures (TLSs) and
germinal centers (GCs), and stratify patient risk, we evaluate external
CD3 recovery, single-cell classification, dense nuclear segmentation,
TLS/GC segmentation, and patient survival prediction.
Across the benchmark, foundation-backed MIPHEI-ViT leads among the
generators, while non-foundation models show task-specific strengths.
Comparing the two routes reveals that internal encoder features
generally outperform generated outputs, and larger output marker panels
do not guarantee stronger downstream utility across datasets.
Moreover, neither internal encoder features nor generated outputs
consistently surpass frozen UNI2.
These findings show that pixel-level stain fidelity does not equate to
downstream utility, and that generators trained at current paired-data
scales have yet to surpass large-scale pure-H\&E foundation models.
With the expansion of paired translation cohorts and the advancement of
foundation-based generative architectures, HEmIF-Bench establishes a definitive benchmark for tracking when H\&E-to-mIF generation crosses the threshold to deliver
genuine value beyond routine histology for tumor microenvironment
analysis and precision oncology.

\end{abstract}


\section{Introduction}
\label{sec:introduction}

Multiplex immunofluorescence (mIF) enables
simultaneous characterization of protein expression,
cellular phenotypes, and tissue organization~\cite{goltsev2018cytof,schurch2020neighborhoods,lin2023orion}.
However, wet-lab mIF acquisition remains costly and
technically demanding, limiting availability
across routine pathology cohorts.
In contrast, hematoxylin and eosin (H\&E) slides
are widely available in clinical archives.
Recent H\&E-to-mIF generators seek to bridge this gap by predicting
spatial protein-expression maps directly from routine
histology~\cite{bian2024hemit,histoplexer2025,balezo2026miphei,valanarasu2026gigatime,wu2025rosie},
raising the prospect of extending multiplex protein analysis to H\&E
collections.
\begin{figure}[!t]
\centering
\includegraphics[width=\textwidth]{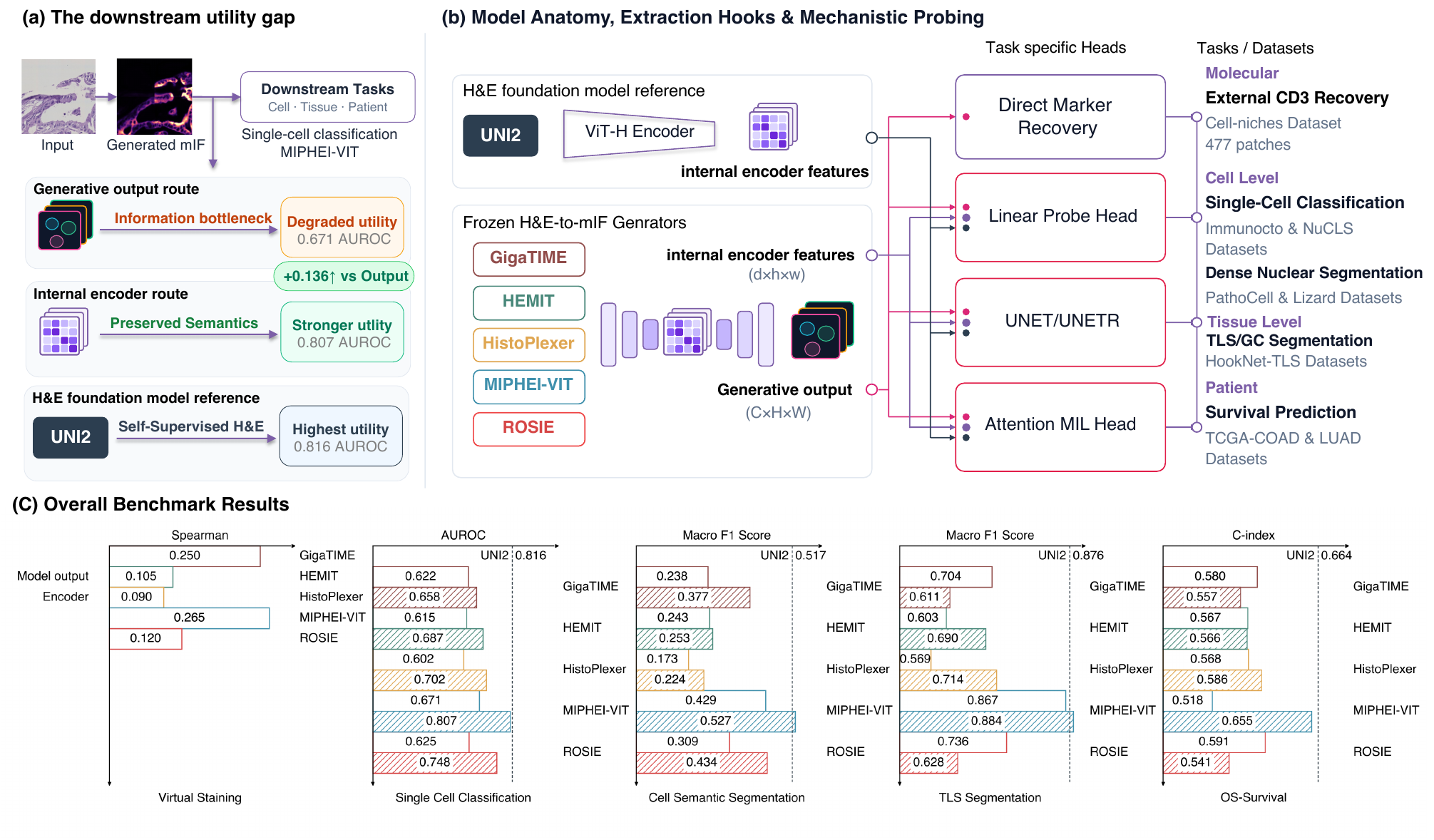}
\caption{\textbf{Overview of HEmIF-Bench.}
Five frozen H\&E-to-mIF generators are evaluated by decoupling
internal encoder features from generated outputs across eight public datasets spanning molecular,
cellular, tissue, and patient scales, with frozen UNI2 serving
as a pure-H\&E pathology foundation reference.}
\label{fig:overview}
\end{figure}

Yet reconstructing plausible virtual stains does not
necessarily establish downstream utility.
Existing virtual staining studies primarily
evaluate pixel-level correlation with measured protein
signals, while recent benchmarks examine only
selected downstream applications~\cite{balezo2026benchmark,klockner2025virtual,sengupta2025utility}.
Existing H\&E-to-mIF generators map input histology through an internal
encoder before decoding generated outputs, whereas self-supervised H\&E
pathology foundation models such as UNI2~\cite{chen2024uni} learn
general-purpose representations directly from routine slides.
Evaluating generated outputs alone leaves untested whether mIF
translation yields stronger internal encoder features than pure-H\&E
foundation models, and whether the generated outputs preserve the
predictive signal of internal encoder features.

To answer these questions, we present \textbf{HEmIF-Bench}, a multi-scale
benchmark evaluating five frozen H\&E-to-mIF generators~\cite{bian2024hemit,histoplexer2025,balezo2026miphei,valanarasu2026gigatime,wu2025rosie}
by decoupling their internal encoder features from generated outputs
and benchmarking both against frozen UNI2~\cite{chen2024uni} (Fig.~\ref{fig:overview}).
Beyond pixel reconstruction, HEmIF-Bench mirrors the clinical workflow
where pathologists rely on mIF to quantify protein signals, resolve
cell phenotypes and nuclear boundaries, delineate tertiary lymphoid
structures (TLSs) and germinal centers (GCs), and stratify patient
risk~\cite{schurch2020neighborhoods,lin2023orion,vanrijthoven2023hooknettls,schallenberg2025cellniches}.
We evaluate all models on five tasks across eight cohorts, including
external CD3 recovery on Cell-niches, single-cell classification on
Immunocto and NuCLS, dense nuclear segmentation on PathoCell and Lizard,
TLS/GC segmentation on HookNet-TLS, and patient survival prediction on
TCGA-COAD and TCGA-LUAD.

Across the benchmark, foundation-backed MIPHEI-ViT leads among the
generators in single-cell classification ($0.807$ macro-AUROC), dense
nuclear segmentation ($0.527$ macro-F1), TLS/GC segmentation ($0.884$
macro-F1), and patient survival ($0.655$ C-index) for internal encoder
features as well as external CD3 recovery ($0.265$ Spearman $\rho$) for
generated outputs, while the four non-foundation models show
task-specific strengths.
Comparing the two routes reveals that generated outputs trail internal
encoder features by $0.093$ mean macro-AUROC ($0.721$ to $0.627$) in
single-cell classification and $0.084$ mean macro-F1 ($0.363$ to
$0.279$) in dense nuclear segmentation, with smaller mean differences in
TLS/GC segmentation ($0.010$ macro-F1, $0.705$ to $0.696$) and patient
survival ($0.017$ C-index, $0.581$ to $0.565$), where GigaTIME and ROSIE
outputs exceed their encoders.
Larger output marker panels do not guarantee stronger downstream
utility, as 16-marker MIPHEI-ViT achieves stronger overall performance
than 50-marker ROSIE and 23-marker GigaTIME.
Moreover, no generator consistently outperforms frozen UNI2 ($0.816$
macro-AUROC, $0.517$ macro-F1, $0.876$ macro-F1, and $0.664$ C-index),
as only MIPHEI-ViT internal encoder features match UNI2 while the other
four encoders and all generated outputs lag.

These findings show that pixel-level stain fidelity does not equate to
downstream utility, and that generators trained at current paired-data
scales have yet to surpass large-scale pure-H\&E foundation models.
As the field scales paired translation cohorts and foundation-based
architectures, HEmIF-Bench establishes a definitive benchmark for
tracking when H\&E-to-mIF generation crosses the threshold to deliver
genuine value beyond routine histology for tumor microenvironment
analysis and precision oncology.

\section{HEmIF-Bench}
\label{sec:method}

\runin{Evaluation protocol.}
HEmIF-Bench evaluates the downstream utility of five frozen
H\&E-to-mIF generators across two routes:
their internal encoder features and generated outputs.
We compare the two routes within each generator and
benchmark both against frozen UNI2~\cite{chen2024uni},
a pure-H\&E pathology foundation reference. External CD3
recovery is assessed separately to distinguish pixel-level
correlation from downstream utility.

For an H\&E patch $x$, let
$g(x)\in\mathbb{R}^{C\times H\times W}$ denote the generated
outputs and $h_g(x)$ the internal encoder features extracted from the
deepest pre-decoder bottleneck layer (or final backbone tokens prior
to decoding). We define:
\begin{align}
\Phi^{\mathrm{enc}}(x)
&= h_g(x)\in\mathbb{R}^{d\times h\times w},
\label{eq:enc-route}\\
\Phi^{\mathrm{out}}(x)
&= g(x)\in\mathbb{R}^{C\times H\times W},
\label{eq:out-route}\\
\phi(x)
&= \frac{1}{|\Omega|}
\sum_{(i,j)\in\Omega}\Phi(x)_{:,i,j}
\in\mathbb{R}^{k},
\label{eq:gap}
\end{align}
where $\Phi\in\{\Phi^{\mathrm{enc}},\Phi^{\mathrm{out}}\}$ denotes either route, with
$(k,|\Omega|)=(d,hw)$ for internal encoder features and
$(C,HW)$ for generated outputs.
Patch-level tasks use globally averaged vectors $\phi(x)$,
whereas dense tasks retain spatial feature maps $\Phi(x)$.

All generators and UNI2 remain frozen, with only downstream
task heads trained. Both routes are evaluated under identical
data partitions, metrics, and random seeds, ensuring that
encoder-to-output performance differences directly reflect
relative downstream utility.

\runin{Tasks.}
HEmIF-Bench mirrors the clinical workflow where pathologists rely on
mIF to quantify protein signals, resolve cell phenotypes and nuclear
boundaries, delineate TLSs and GCs, and stratify patient
risk~\cite{schurch2020neighborhoods,lin2023orion,vanrijthoven2023hooknettls,schallenberg2025cellniches}.
At the molecular scale, \textbf{external CD3 recovery} measures how
accurately generators reconstruct target protein signals on unseen
tissue.
At the cellular scale, \textbf{single-cell classification} and
\textbf{dense nuclear segmentation} test whether models can distinguish
immune and stromal cell types with similar H\&E morphology and separate
crowded nuclei.
At the tissue scale, \textbf{TLS/GC segmentation} tests the delineation
of TLSs and GCs defined by B- and T-cell spatial
co-localization~\cite{vanrijthoven2023hooknettls}.
At the patient scale, \textbf{patient survival prediction} tests
whether WSI-level features stratify overall survival on clinical
cohorts without wet-lab mIF.

\runin{Datasets.}
We evaluate these five tasks across eight public datasets
(Table~\ref{tab:datasets}).

\begin{table}[H]
\caption{Benchmark tasks and datasets across four analytical scales.
GT: ground-truth annotation source; FOVs: fields of view; ROIs:
regions of interest; WSIs: whole-slide images.}
\label{tab:datasets}
\centering
\footnotesize
\setlength{\tabcolsep}{4pt}
\resizebox{\textwidth}{!}{%
\begin{tabular}{@{}llllll@{}}
\toprule
Scale & Task & Dataset & Organ & GT source & Format \& Data scale \\
\midrule
Molecular & External CD3 recovery
& Cell-niches~\cite{schallenberg2025cellniches}
& Lung & Paired mIF (CD3)
& $477$ patches, $20$ FOVs \\
\midrule
\multirow{4}{*}{Cellular}
& \multirow{2}{*}{Single-cell classification}
& Immunocto~\cite{simard2025immunocto}
& Colorectal & mIF-defined subtypes
& $2482157$ patches \\
& & NuCLS~\cite{amgad2022nucls}
& Breast & Pathologist H\&E
& $59374$ patches / $1744$ Fovs \\
\cmidrule(l){2-6}
& \multirow{2}{*}{Dense nuclear segmentation}
& PathoCell~\cite{luscher2025pathocellbench}
& Colorectal & mIF-guided nuclei
& $109$ Fovs \\
& & Lizard~\cite{graham2021lizard}
& Colorectal & Pathologist H\&E
& $238$ Fovs \\
\midrule
Tissue & TLS/GC segmentation
& HookNet-TLS~\cite{vanrijthoven2023hooknettls}
& Lung & IHC-guided TLS/GC
& $32306$ patches / $767$ WSIs \\
\midrule
\multirow{2}{*}{Patient}
& \multirow{2}{*}{Patient survival prediction}
& TCGA-COAD~\cite{tomczak2015tcga}
& Colorectal & Clinical follow-up
& $458$ WSIs / $450$ patients \\
& & TCGA-LUAD~\cite{hoadley2018cell}
& Lung & Clinical follow-up
& $541$ WSIs / $478$ patients\\
\bottomrule
\end{tabular}%
}
\end{table}

For external CD3 recovery, we use paired H\&E and mIF lung cancer
cores from Cell-niches~\cite{schallenberg2025cellniches}, targeting
CD3 as the sole marker shared across all five generators.
For single-cell classification, we pair Immunocto~\cite{simard2025immunocto},
a colorectal cohort with mIF-defined immune subtypes, and
NuCLS~\cite{amgad2022nucls}, a breast cancer cohort with pathologist
H\&E annotations, to evaluate both mIF-guided and cross-organ
phenotyping.
For dense nuclear segmentation, we similarly pair mIF-annotated
PathoCell~\cite{luscher2025pathocellbench} with pathologist-annotated
colorectal H\&E images in Lizard~\cite{graham2021lizard}.
For TLS/GC segmentation, we use lung cancer WSIs with
immunohistochemistry-guided TLS and GC annotations from
HookNet-TLS~\cite{vanrijthoven2023hooknettls}.
For patient survival prediction, we use colorectal (TCGA-COAD) and
lung adenocarcinoma (TCGA-LUAD) WSIs~\cite{tomczak2015tcga,hoadley2018cell},
matching the primary training organs of the evaluated generators
(Table~\ref{tab:Methods}).

\runin{Benchmark methods.}
We evaluate five publicly released H\&E-to-mIF generators alongside the
pathology foundation reference UNI2~\cite{chen2024uni}
(Table~\ref{tab:Methods}).
Four generators are trained without pathology foundation pretraining,
including HEMIT~\cite{bian2024hemit} with a dual-branch CNN-Swin
pix2pix architecture, HistoPlexer~\cite{histoplexer2025} with a
conditional GAN, GigaTIME~\cite{valanarasu2026gigatime} with a
NestedUNet, and ROSIE~\cite{wu2025rosie} with a ConvNeXt-S backbone.
In contrast, MIPHEI-ViT~\cite{balezo2026miphei} adapts the frozen
pathology foundation model H-optimus-0~\cite{hoptimus0} via low-rank
adaptation (LoRA) within a U-Net.
Across the five generators, internal encoder features span $d\in[256,
1536]$ and generated outputs span $C\in[3, 50]$ markers across
colorectal, lung, melanoma, and multi-organ cohorts.

\begin{table}[H]
\caption{Benchmark methods. $d$: encoder width;
$C$: marker channels. Native tile: pixels@resolution
($\mu$m/px), with physical extent.
UNI2 and MIPHEI-ViT use pathology foundation backbones.
GigaTIME alone was trained on the same organ and subtype
(lung adenocarcinoma) as Cell-niches. $^\dagger$Encoder-only reference.}
\label{tab:Methods}
\centering
\footnotesize
\setlength{\tabcolsep}{4pt}
\resizebox{\textwidth}{!}{%
\begin{tabular}{@{}llrrll@{}}
\toprule
Model & Backbone & $d$ & $C$ & Native tile & Training cohort \\
\midrule
HEMIT~\cite{bian2024hemit}
& dual-branch pix2pix (CNN\,+\,Swin)
& 512 & 3
& $1024$@$0.25$ ($256\mu$m)
& colon mIHC, 5{,}292 pairs \\
HistoPlexer~\cite{histoplexer2025}
& cGAN
& 256 & 11
& $1024$@$0.25$ ($256\mu$m)
& melanoma, IMC, 78 patients \\
MIPHEI-ViT~\cite{balezo2026miphei}
& U-Net\,+\,H-optimus-0~\cite{hoptimus0} (LoRA)
& 1536 & 16
& $256$@$0.5$ ($128\mu$m)
& Orion colorectal~\cite{lin2023orion}, mIF \\
GigaTIME~\cite{valanarasu2026gigatime}
& NestedUNet
& 512 & 23
& $256$@$0.5$ ($128\mu$m)
& lung adenocarcinoma, mIF, 21 patients \\
ROSIE~\cite{wu2025rosie}
& ConvNeXt-S
& 768 & 50
& $128$@$0.25$ ($32\mu$m)
& 18 studies (tonsil-rich), CODEX \\
\midrule
UNI2$^{\dagger}$~\cite{chen2024uni}
& ViT-H
& 1536 & ---
& $224$@$0.5$ ($112\mu$m)
& pan-cancer WSIs \\
\bottomrule
\end{tabular}%
}
\end{table}

\runin{Implementation details.}
Each frozen model processes inputs at its native tile resolution and
physical extent (Table~\ref{tab:Methods}). ROSIE predicts channel-wise
mean expression over local $8\times 8$-pixel regions rather than
pixel-wise expression. Its local predictions are spatially aggregated
for external CD3 recovery and dense segmentation tasks.
For external CD3 recovery on Cell-niches, predicted and ground-truth
CD3 maps are mean-pooled to an $8\,\mu\mathrm{m}$ grid and evaluated
via patch-level Spearman $\rho$ and area under the precision--recall
curve (AUPRC) against per-patch Otsu-thresholded ground-truth masks,
with 95\% confidence intervals (CIs) estimated via $2{,}000$-resample
FOV bootstrap.
For single-cell classification on Immunocto~\cite{simard2025immunocto} and
NuCLS~\cite{amgad2022nucls}, pooled cell-patch vectors $\phi(x)$ are
evaluated via a shared linear probe in macro-AUROC. Because Immunocto is cropped from 40 Orion colorectal
slides~\cite{lin2023orion} that were also used to train
MIPHEI-ViT~\cite{balezo2026miphei}, our 32/4/4 train/validation/test
slide split excludes all MIPHEI-ViT training slides from testing (with
two test slides overlapping its validation set) and subsamples the
dominant \emph{other} class to match the CD4 count within each split.
For dense nuclear segmentation (PathoCell, Lizard) and TLS/GC
segmentation (HookNet-TLS), spatial feature maps $\Phi^{\mathrm{enc}}(x)$
and $\Phi^{\mathrm{out}}(x)$ are decoded via route-matched lightweight
upsampling and convolutional segmentation heads and evaluated using
macro-F1.
For patient survival prediction on TCGA-COAD and TCGA-LUAD, WSI
patch-level vectors $\phi(x)$ are aggregated into patient risk scores
via five-fold cross-validated attention multiple-instance
learning~\cite{lu2021wsi,lei2025shazam} and evaluated using the
concordance index (C-index).

To test the effects of feature dimensionality ($d\gg C$;
Table~\ref{tab:Methods}) and pathology foundation pretraining under the
single-cell classification linear-probe protocol on Immunocto and NuCLS,
we include two controlled comparisons.
First, we standardize pooled internal encoder features and project them
via training-fitted PCA to each generator's marker count $C$, denoted as
$\mathrm{enc}@C$.
Second, we evaluate frozen H-optimus-0~\cite{hoptimus0} alongside
MIPHEI-ViT, testing paired differences via case-clustered bootstrap 95\%
CIs.


\section{Experiments and Results}
\label{sec:results}

\begin{figure}[H]
\centering
\includegraphics[width=\textwidth]{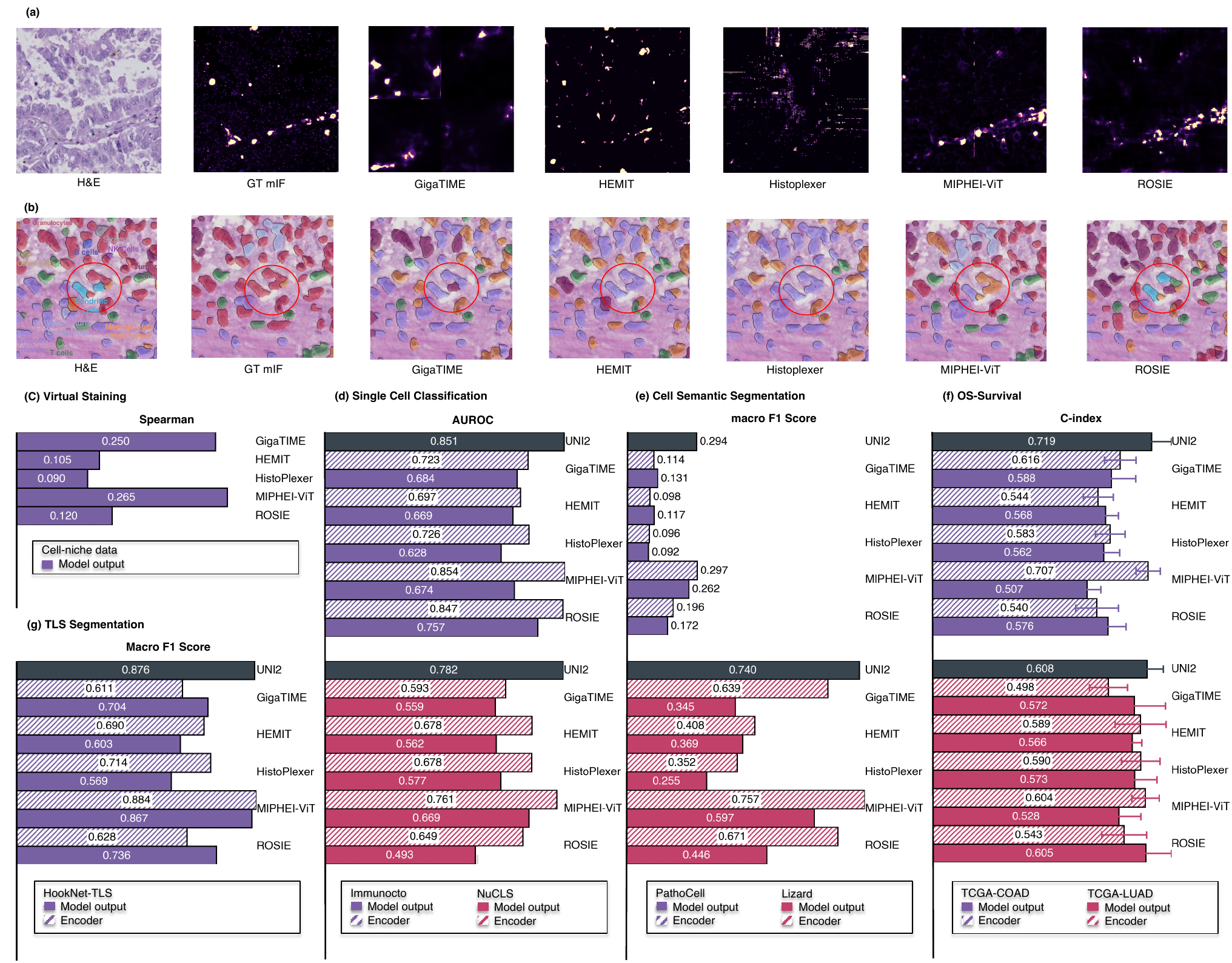}
\caption{\textbf{Main benchmark results.}
\textbf{(a)}--\textbf{(b)}~Dense nuclear segmentation on a
PathoCell example:
\textbf{(a)}~H\&E, ground-truth mIF, and generated outputs;
\textbf{(b)}~predicted segmentation maps. The red circle marks
a rare cell correctly segmented only by ROSIE in this example.
\textbf{(c)}~External CD3 recovery on Cell-niches ($20$ FOVs):
Spearman $\rho$ after mean pooling to $8\,\mu\mathrm{m}$.
Displayed CD3 examples use rank-preserving histogram
matching for visualization only.
\textbf{(d)}~Single-cell classification: macro-AUROC on
Immunocto and NuCLS.
\textbf{(e)}~Dense nuclear segmentation: macro-F1 on
PathoCell and Lizard.
\textbf{(g)}~TLS/GC segmentation: macro-F1 on HookNet-TLS.
\textbf{(f)}~Patient survival prediction: C-index on TCGA-COAD
and TCGA-LUAD (five-fold mean $\pm$ fold s.d.\
where shown).}
\label{fig:results}
\end{figure}

\runin{MIPHEI-ViT leads among generators, while non-foundation models
show task-specific strengths.}
Across the five H\&E-to-mIF generators in Fig.~\ref{fig:results}, MIPHEI-ViT~\cite{balezo2026miphei} achieves the
strongest overall performance.
Its internal encoder features rank first among the five generators on
all seven downstream datasets, reaching $0.854$ and $0.761$ macro-AUROC
on Immunocto and NuCLS, $0.297$ and $0.757$ macro-F1 on PathoCell and
Lizard, $0.884$ macro-F1 on HookNet-TLS, and $0.707$ and $0.604$
C-index on TCGA-COAD and TCGA-LUAD, while its generated outputs rank
first in external CD3 recovery at $0.265$ 

Spearman $\rho$ and on NuCLS
at $0.669$, PathoCell at $0.262$, Lizard at $0.597$, and HookNet-TLS at
$0.867$.
This lead reflects its pathology foundation backbone
H-optimus-0~\cite{hoptimus0} and wide $1536$-dimensional feature
channels in Table~\ref{tab:Methods} compared with the four generators
trained without foundation pretraining.

Among the four non-foundation generators, no single model dominates
across datasets, as each exhibits strengths tied to its training cohort,
marker panel, and native tile size in Table~\ref{tab:Methods}.
Trained across 18 CODEX studies on $32\,\mu\mathrm{m}$ crops with $50$
markers, ROSIE~\cite{wu2025rosie} ranks first among all generated
outputs on mIF-defined Immunocto at $0.757$ macro-AUROC and TCGA-LUAD
survival at $0.605$ C-index, while ranking second overall in dense
nuclear segmentation on PathoCell and Lizard with $0.196$ and $0.671$
macro-F1 for internal encoder features and $0.172$ and $0.446$ macro-F1
for generated outputs in Fig.~\ref{fig:results}a--b.
Lung-trained GigaTIME~\cite{valanarasu2026gigatime} ranks second in
external lung CD3 recovery at $0.250$ Spearman $\rho$ and achieves the
highest C-index among generated outputs on TCGA-COAD at $0.588$.

In contrast, HEMIT~\cite{bian2024hemit} and
HistoPlexer~\cite{histoplexer2025}, which process $1024\times 1024$
tiles covering a $256\,\mu\mathrm{m}$ physical extent, generalize better
to cross-organ pathologist H\&E annotations on NuCLS with $0.678$
macro-AUROC for internal encoder features and $0.562$ and $0.577$
macro-AUROC for generated outputs, outperforming GigaTIME and ROSIE
while also achieving the top non-foundation scores for internal encoder
features at $0.690$ and $0.714$ on HookNet-TLS and $0.589$ and $0.590$
on TCGA-LUAD.

\runin{Internal encoder features outperform generated outputs.}
Comparing the two routes within each generator in
Fig.~\ref{fig:results}d--g shows that internal encoder features
outperform generated outputs in $27$ of $35$ model--dataset evaluations.
Across all five generators, generated outputs consistently trail
internal encoder features on Immunocto by $0.028$ to $0.180$
macro-AUROC, on NuCLS by $0.034$ to $0.156$ macro-AUROC, and on Lizard
by $0.039$ to $0.293$ macro-F1.

\begin{figure}[!t]
\centering
\includegraphics[width=\textwidth]{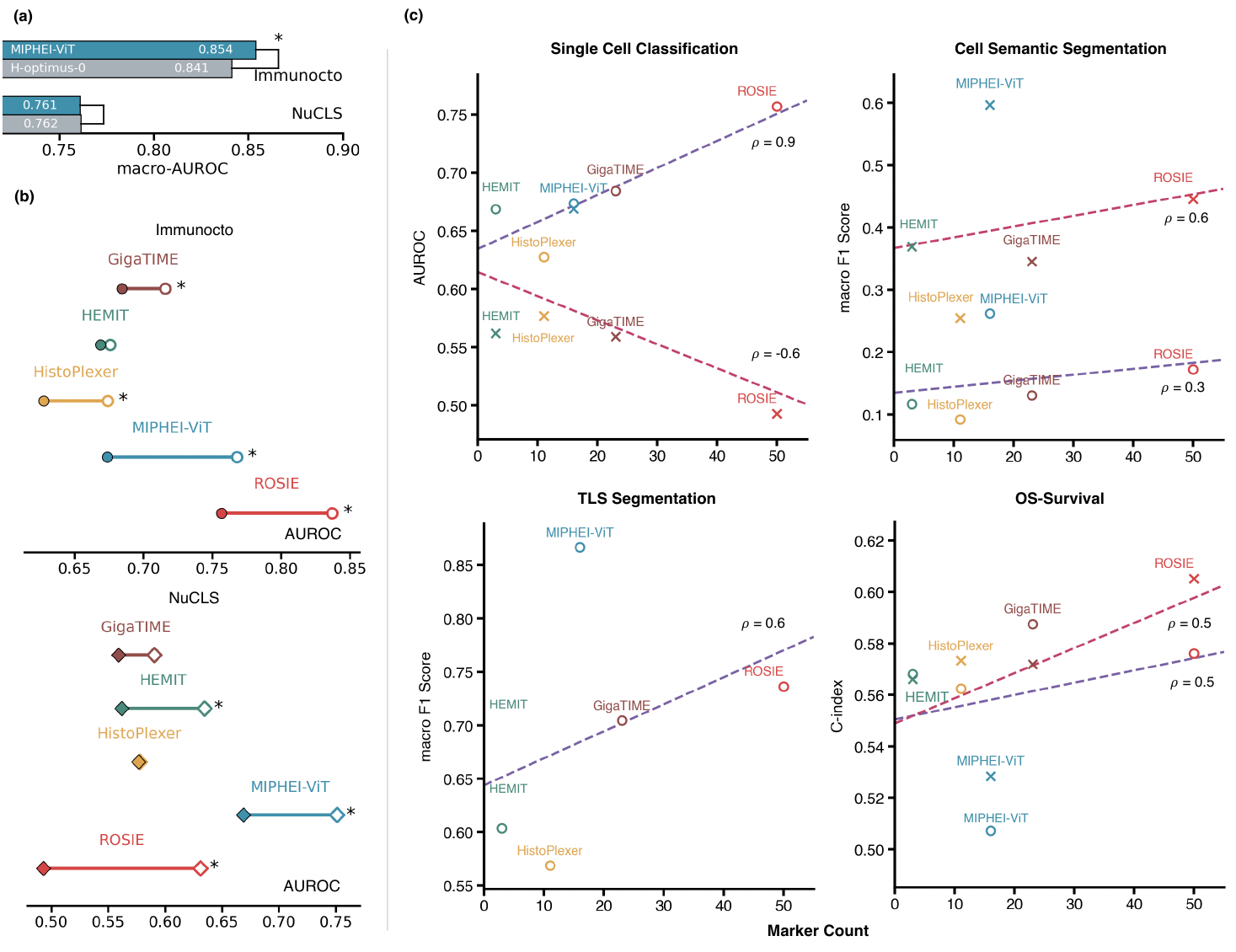}
\caption{\textbf{Controlled analyses of internal encoder features and
generated outputs.}
\textbf{(a)}~Foundation backbone control: macro-AUROC for frozen
H-optimus-0 versus MIPHEI-ViT internal encoder features on Immunocto
and NuCLS.
\textbf{(b)}~Dimensionality control: $\mathrm{enc}@C$ (filled markers)
versus generated outputs ($\Phi^{\mathrm{out}}$, open markers) on
Immunocto and NuCLS.
\textbf{(c)}~Output panel breadth: downstream performance of generated
outputs across marker counts $C\in[3, 50]$, with least-squares trend
lines and Spearman $\rho$.
In \textbf{(a)}--\textbf{(b)}, $^{*}$ denotes a paired case-clustered
bootstrap 95\% CI excluding zero.}
\label{fig:controls}
\end{figure}

Route differences are mixed on PathoCell, HookNet-TLS, TCGA-COAD, and
TCGA-LUAD, where generated outputs from GigaTIME and ROSIE exceed their
weaker internal encoder features by $0.093$ and $0.109$ macro-F1 on
HookNet-TLS and by $0.074$ and $0.062$ C-index on TCGA-LUAD, whereas
the top-performing internal encoder features of MIPHEI-ViT drop after
decoding on every dataset, losing $0.180$ macro-AUROC on Immunocto,
$0.160$ macro-F1 on Lizard, and $0.200$ C-index on TCGA-COAD.

To test whether this performance drop is merely an artifact of reduced
channel dimensionality $C\ll d$ in Table~\ref{tab:Methods}, we project
internal encoder features via PCA to the generator's output channel
count $C$ as $\mathrm{enc}@C$ in Fig.~\ref{fig:controls}b, which still
outperforms generated outputs across all five generators on both
Immunocto and NuCLS.

\runin{Larger output marker panels do not guarantee stronger downstream
utility.}
Evaluating generated outputs across panel breadth $C\in[3, 50]$ in
Fig.~\ref{fig:controls}c shows that predicting more protein channels
improves downstream performance only on cohort-aligned Immunocto, where
macro-AUROC increases with marker count at Spearman $\rho=0.900$ as
50-marker ROSIE reaches $0.757$ and 23-marker GigaTIME reaches $0.684$.
Across the other six downstream datasets and external CD3 recovery,
generated outputs show no statistically significant positive correlation
with marker count.

On cross-organ NuCLS, performance trends negatively with panel breadth
at $\rho=-0.600$, as 3-marker HEMIT at $0.562$ and 11-marker HistoPlexer
at $0.577$ outperform both 23-marker GigaTIME at $0.559$ and 50-marker
ROSIE at $0.493$.
Similarly, 16-marker MIPHEI-ViT outperforms both 23-marker GigaTIME and
50-marker ROSIE in generated outputs on NuCLS, PathoCell, Lizard,
HookNet-TLS, and Cell-niches CD3 recovery, indicating that upstream
backbone quality and cohort alignment outweigh raw channel count.

\runin{H\&E-to-mIF generators do not surpass the pure-H\&E foundation
model UNI2.}
As the pure-H\&E reference in Fig.~\ref{fig:results}d--g, frozen
UNI2~\cite{chen2024uni} reaches $0.851$ and $0.782$ macro-AUROC on
Immunocto and NuCLS, $0.294$ and $0.740$ macro-F1 on PathoCell and
Lizard, $0.876$ macro-F1 on HookNet-TLS, and $0.719$ and $0.608$
C-index on TCGA-COAD and TCGA-LUAD.

Against this baseline, generated outputs from all five generators fall
short across all $35$ evaluations by $0.003$ to $0.485$, while internal
encoder features from the four non-foundation generators trail across
all $28$ evaluations by $0.004$ to $0.388$ and MIPHEI-ViT remains within
$-0.021$ to $+0.018$ of UNI2 across the seven datasets.

Comparing MIPHEI-ViT against its frozen backbone
H-optimus-0~\cite{hoptimus0} on single-cell classification in
Fig.~\ref{fig:controls}a shows that mIF adaptation improves internal
encoder features on cohort-aligned Immunocto, whereas the difference on
cross-organ NuCLS is not statistically significant.


\FloatBarrier
\section{Discussion and Conclusion}
\label{sec:discussion}

Virtual mIF generation is moving beyond image translation toward
molecularly informed analysis of routine
histology~\cite{balezo2026benchmark,sengupta2025utility}, yet pixel-level
marker reconstruction alone does not establish downstream utility.
HEmIF-Bench demonstrates the necessity of disentangling external CD3
recovery from the downstream utility of internal encoder features and
generated outputs, as these evaluations yield distinct rankings across
the five generators.

Our findings highlight two key challenges for building molecularly
informed pathology models.
First, given the limited scale of current paired H\&E--mIF datasets
compared with massive pan-cancer H\&E pretraining corpora
(Table~\ref{tab:Methods}), mIF translation on existing cohorts improves
internal encoder features only on cohort-aligned data (e.g., MIPHEI-ViT
over frozen H-optimus-0 on Immunocto) and does not consistently
outperform the pure-H\&E pathology foundation reference
UNI2~\cite{chen2024uni} across cellular, tissue, and patient tasks.
Second, strong internal encoder features do not automatically yield
equally useful generated outputs, as generated outputs trail internal
encoder features across most tasks and this drop persists even after
matching feature dimensionality ($\mathrm{enc}@C$) in single-cell
classification.

Our study has two limitations.
First, external marker recovery is evaluated only on CD3 in lung tissue,
as CD3 is the sole marker shared across all five generators; future work
can extend this evaluation to additional markers and organs as broader
paired H\&E--mIF datasets become available.
Second, we evaluate all generators in a frozen setting and test internal
encoder features and generated outputs separately; exploring task-specific
fine-tuning or combining the two routes remains an interesting direction
for future work.

HEmIF-Bench evaluates five H\&E-to-mIF generators against the pathology
foundation reference UNI2 across external CD3 recovery and cellular,
tissue, and patient-level tasks.
Across the benchmark, foundation-backed MIPHEI-ViT leads among the
generators, while non-foundation models show task-specific strengths.
Comparing the two routes reveals that internal encoder features
generally outperform generated outputs, and larger output marker panels
do not guarantee stronger downstream utility across datasets.
Neither internal encoder features nor generated outputs consistently
surpass frozen UNI2.
As the field scales paired H\&E--mIF cohorts for foundation-based
generation and joint histology--omics representation
learning~\cite{li2026hex,chen2025omiclip}, HEmIF-Bench establishes a
definitive benchmark for tracking when H\&E-to-mIF generation crosses
the threshold to deliver genuine biological and clinical value beyond
routine histology, ultimately guiding the translation of routine H\&E
archives into scalable, molecularly informed tools for tumor
microenvironment analysis and precision oncology.

\begin{credits}
\subsubsection{\ackname}
This work was supported by the Research Grants Council of Hong Kong
(Grants 16103224, 16103625, 16102322, 16100926, and N\_HKUST648/24 to H.P.).

\subsubsection{\discintname}
The authors have no competing interests to declare that are relevant
to the content of this article.
\end{credits}

\bibliographystyle{splncs04}
\bibliography{references}

\end{document}